\documentclass[letterpaper]{article} 
\usepackage[final]{aaai25}
\usepackage{times}  
\usepackage{helvet}  
\usepackage{courier}  
\usepackage[hyphens]{url}  
\usepackage{graphicx} 
\usepackage{natbib}  
\usepackage{caption} 
\usepackage{algorithm}
\usepackage{algorithmic}

\usepackage{graphicx}
\usepackage{multirow}
\usepackage{amsmath}
\usepackage{array}
\usepackage{booktabs}
\usepackage{xcolor}

\usepackage{newfloat}
\usepackage{listings}
\DeclareCaptionStyle{ruled}{labelfont=normalfont,labelsep=colon,strut=off} 
\floatstyle{ruled}
\newfloat{listing}{tb}{lst}{}
\floatname{listing}{Listing}
\title{Improving Physics Reasoning in Large Language Models Using Mixture of Refinement Agents}

\author {
    Raj Jaiswal ,
    Dhruv Jain ,
    Harsh Parimal Popat ,
    Avinash Anand ,
    Abhishek Dharmadhikari ,
    Atharva Marathe ,
    Rajiv Ratn Shah 
}
\affiliations {
    \{jaiswalp, harsh21048, avinasha, rajivratn\}@iiitd.ac.in, dhruv.jain.ece21@itbhu.ac.in,\\
    \{abhishekdharmadhikari25, atharvamarathe8\}@gmail.com
    
}
\usepackage{bibentry}

\begin{document}

\maketitle

\begin{abstract}
Large Language Models (LLMs) demonstrate remarkable capabilities in various reasoning tasks. However, they encounter significant challenges when it comes to scientific reasoning, particularly in physics, which requires not only mathematical reasoning but also factual and conceptual understanding. When addressing complex physics problems, LLMs typically face three key issues: problem miscomprehension, incorrect concept application, and computational errors. While each of these problems can be addressed individually, there is a need for a generalized approach that can tackle all three issues simultaneously. To address this, we introduce Mixture of Refinement Agents (MoRA), a novel agentic refinement framework that iteratively refines the LLM generated base solution by correcting the aforementioned errors, resulting in a significant performance improvement for open-source LLMs. Our approach aims to bridge the gap between open-source LLMs and GPT-4o by utilizing the latter as error identifier to guide these refinement agents. We evaluate our approach on the SciEval and MMLU subsets along with our own physics dataset (PhysicsQA). MoRA significantly improves the performance of Llama-3-70B and Gemma-2-27B on these datasets, achieving up to a 16\% increase in final answer accuracy.
\end{abstract}

\section{Introduction}

Scientific reasoning, particularly in field of physics, requires a deep understanding that spans multiple disciplines. It demands not only domain-specific knowledge but also the integration of mathematical reasoning with theoretical concepts, applying abstract principles and formulae across various contexts and scenario. Successfully solving these challenges is a fundamental aspect of human intelligence, as it entails not just recalling information but adapting knowledge to solve diverse complex problems.

\begin{figure*}[ht]
    \centering
    \includegraphics[width=15cm]{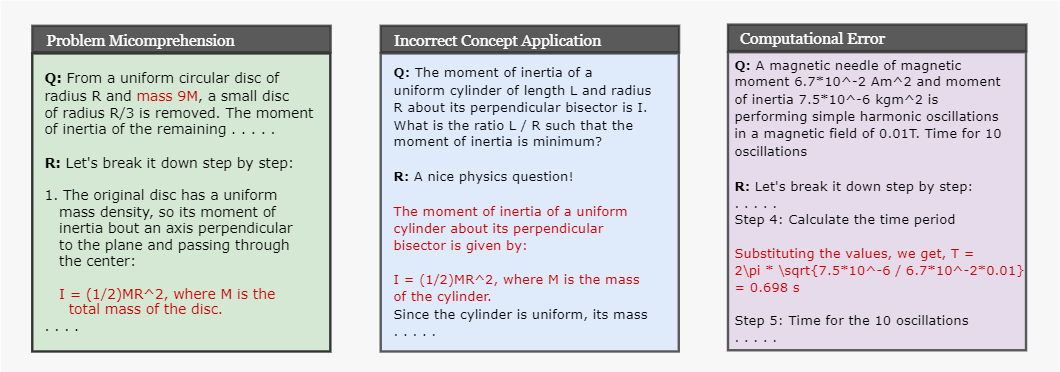} 
    \caption{The illustration of three key error observations in the CoT solution of open source LLMs for physics problems. (a) showcases problem miscomprehension, where the LLM response uses the incorrect value of variables given in the question here, M instead of 9M, (b) showcases incorrect concept application in the LLM response, here incorrect moment of inertia formula for uniform cylinder, (c) demonstrate computational error within LLM response here, incorrect calculation of time period.}
    \label{fig:error}
\end{figure*}

Solving complex physics problems still remains a challenge for open source LLMs. The difficulty stems from the need to integrate both mathematical and domain-specific knowledge while engaging in multi-hop, step-by-step reasoning. One approach to address this challenge can be collecting question and solution trajectory annotations and finetune LLMs to enhance these capabilities, similar to recent mathematical reasoning works \cite{luo2023wizardmath, yuan2024scaling}. However, the process of such annotations and finetuning is time-consuming and costly. On the other hand, solutions generated by LLMs for physics problems using CoT prompting \cite{wei2022chain} often contain errors, such as objective misalignment, incorrect formula application, and computational mistakes, as illustrated in Figure \ref{fig:error}. Moreover, solutions to multihop physics problems contain multiple such errors together. 

Open source LLMs struggles to accurately directly identify reasoning mistakes in their own solutions \cite{li2024evaluating, tyen-etal-2024-llms,anand2023context}, making them unreliable for error detection and self-refinement. While objective alignment errors can be corrected once identified, refining computational and conceptual errors requires strong mathematical reasoning and contextual understanding of the specific question. Addressing all these different errors simultaneously remains a significant challenge for open-source LLMs.

This motivated us to develop the Mixture of Refinement Agents (MoRA) framework. MoRA iteratively refines LLM responses through a two-step process in each iteration. First, the framework leverages a advanced model to identify various errors within the solution using appropriate flags and scores. In the next step, based on the identified errors, prioritized agent routing is conducted, in which the appropriate agents are activated to address and mitigate the specific errors. This process results in a progressively refined solution.

In the domain of physics, evaluation benchmarks are essential for assessing the conceptual and mathematical reasoning of LLMs. Benchmarks like MMLU, SciEval \cite{sun2024scieval}, and ScienceQA \cite{lu2022learn} focus on foundational knowledge and general reasoning, while more challenging ones like OlympiadBench \cite{he2024olympiadbench} and JEEBench \cite{arora2023have} require advanced reasoning skills. To bridge the gap, we curated our own dataset PhysicsQA, containing set of diverse, intermediate-level high school physics problems that provide a balanced challenge, allowing a exhaustive evaluation and analysis of open-source LLMs on physics problems.

We perform exhaustive experimentation of MoRA across four datasets including PhysicsQA as shown in Table \ref{experimentalresults2}. MoRA improves accuracy on the PhysicsQA benchmark over CoT-generated solutions by 13.38\% for Llama-3-70B and by 16.03\% for Gemma-2-27B. This significant enhancement highlights MoRA’s effectiveness in refining solutions, particularly in complex and diverse physics problems as in PhysicsQA. Our further analysis offers insights into the error distribution across different models and evaluates the effectiveness of individual refinement agents based on their refinement rates.



\section{Related Works}

\subsubsection{LLM Reasoning}
LLMs have been successfully applied to address multi-step reasoning tasks by generating intermediate reasoning steps, referred to as Chain-of-Thought (CoT) \cite{wei2022chain}, Auto-CoT \cite{zhang2022automatic}, and Complex-CoT \cite{fu2022complexity}, among others. Advanced techniques like Iter-CoT \cite{sun2023enhancing} and ToT \cite{yao2024tree} extend these capabilities but remain constrained by the knowledge in training data and the specific structures they were designed with. While In-Context Learning (ICL) \cite{brown2020language} has significantly improved LLM performance, challenges like hallucinations and limitations in reasoning flexibility persist.

\subsubsection{LLMs for Scientific Reasoning}
LLMs face significant limitations in complex knowledge reasoning tasks \cite{petroni2020kilt}. \cite{ouyang2023structured} introduced a structured reasoning strategy to guide LLMs in solving complex chemistry problems, enabling them to generate high-quality reasoning. Solving these problems requires not only domain knowledge, like formulae and calculations, but also a step-by-step reasoning process. \cite{ma2024sciagent} proposed a method where agents generate a high-level plan based on the question, retrieve relevant functions from a toolset, and execute low-level actions by integrating natural language and Python code.

\subsubsection{Self Verification with LLMs}
Recent works \cite{cobbe2021training, ling2024deductive} have attempted to address the challenge of error detection in step-by-step reasoning. However, these methods often require additional training data or domain-specific exemplars, making them less practical. \cite{miao2023selfcheck} proposes using the LLM itself to verify the conditional correctness of each step in the reasoning chain, similar to how a human reviews their work.\cite{anand2023kg} Accurate error recognition and correction are crucial for enhancing problem-solving capabilities, as demonstrated by \cite{li2024evaluating}, which defines tasks to assess LLMs' mathematical reasoning abilities in error identification and correction.

\subsubsection{LLMs for Mathematical Reasoning}
LLMs tends to struggle with arithmetic calculations when solving math problems \cite{cobbe2021training, gao2023pal}. However, incorporating code generation and execution has shown promise in enhancing the accuracy of mathematical reasoning. Leveraging these strengths, the GPT-4 Code Interpreter \cite{zhou2023solving} has been integral to frameworks like MathCoder \cite{wang2023mathcoder}, which is designed to improve the mathematical reasoning capabilities of open-source models. Findings from \cite{zhou2023solving} indicate that GPT-4 Code’s impressive proficiency in solving mathematical problems is largely due to its step-by-step code generation and the dynamic refinement of solutions based on code execution outcomes.

\subsubsection{LLM Reasoning with external database}
\cite{lewis2020retrieval} proposed RAG framework, which incorporates \cite{SciPhyRAG} a retrieval component to fetch relevant information from a given knowledge base. Integrating LLMs with knowledge representation tools, such as knowledge graphs (KGs) \cite{mruthyunjaya2023rethinking}, has further enhanced reasoning capabilities. \cite{yao2024tree} demonstrated that augmenting LLMs with comprehensive external knowledge from KGs can significantly improve their performance and facilitate more robust reasoning processes. A notable example is GraphRAG \cite{edge2024local}, a retrieval enhancement technique that leverages knowledge graphs to map relationships between entities, thereby enhancing the retrieval process using large language models (LLMs).


\section{Dataset: PhysicsQA}

Our dataset comprises 370 carefully selected high school physics questions sourced from online resources. These questions are notably complex, often requiring the application of multiple concepts, intricate computations, \cite{highdatasetavinash} and multihop reasoning. Each question is paired with a comprehensive, step-by-step solution, to support the evaluation \cite{anand2024geovqa} and fine-tuning of LLMs for physics reasoning. Table \ref{physicsqa} illustrates the topic-wise distribution of the questions, providing a clear overview of the areas covered. PhysicsQA offers a more robust evaluation and analysis of LLM performance by encompassing a diverse range of questions, both in terms of complexity and the topics covered.

\begin{table}[ht]
\centering
\begin{tabular}{l|c c}
\hline
\textbf{Chapter Name} & \textbf{Percentage}  \\
\hline
Electromagnetism & 29.8\% \\
Mechanics and Kinematics & 21.8\%  \\
Thermodynamics and Heat & 15.7\% \\
Waves and Optics & 15.4\%  \\
Nuclear and Modern Physics & 8.9\% \\
Material Properties and Elasticity & 8.3\% \\
\bottomrule
\end{tabular}
\caption{Topic-wise Distribution in PhysicsQA}
\label{physicsqa}
\end{table}


\section{Mixture of Refinement Agents}
This section introduces our mixture of refinement agents (MoRA) framework. We first discuss our motivation behind MoRA; then, we introduce the error identification stage and refinement agents. Finally, we discuss how these agents are routed iteratively to correct different errors in the solutions generated by the LLM.

\subsection{Motivation}

While analyzing physics problems and their CoT solutions generated with LLMs (Llama-3-70B \& Gemma-2-27B), we observed three key errors made by them:

\textbf{\textit{Observation 1:}} \textit{LLMs in few cases struggle to fully grasp the objective of the question, along with misinterpreting the values of variables and constants provided in the question.}

Although this issue has been identified in only a few cases, it is significant one because it leads to solutions that fails to address the correct interpretation of a given question resulting in \textit{problem miscomprehension}.

\textbf{\textit{Observation 2:}} \textit{LLMs struggle to apply the correct concepts or formulae with respect to the context of the given problem.}

This issue is a more recurring one in LLMs, especially for problems requiring considering a specific case rather than relying on a generic formula. \cite{anand2024PhyRLHF} For instance, the formula for calculating the \textit{moment of inertia} varies depending on the distribution of mass. 

\textbf{\textit{Observation 3:}} \textit{Many physics problems involve mathematical reasoning and algebraic computation, areas where LLMs tend to struggle.}

Computational errors account for the majority of errors in solutions generated by LLMs. LLMs struggles with accurate algebraic and arithmetic computations resulting in errors within the reasoning and final answer.
 
While these three issues can be addressed individually, solutions often exhibit multiple errors together. Therefore, a single framework is required to rectify all three issues effectively, which motivated us to develop the MoRA. We first perform error identification on a given solution; then these errors are mitigated iteratively using specialized refinement agents, resulting in accurate solutions. 

\subsection{Error Identification}

The errors in the solutions are classified into three categories: 1) \textit{problem miscomprehension}, 2) \textit{incorrect concept application}, and 3) \textit{computational errors} as showed in Figure \ref{fig:error}.

For error identification, we choose to rely on GPT-4o. Our experiments and analysis shows that GPT-4o showcases superior performance compared to other models, particularly in problem comprehension and correct physics concept application required. Thus, it is adequate for locating errors within solutions generated by other models. Given a question and it's LLM response, we prompt GPT-4o to identify and locate different errors in the solution using the combination of following flags and scores:\\

\textbf{Problem Comprehension Flags:} We prompt GPT-4o to check for the problem miscomprehension using the following two flags: (i) \textbf{Objective Alignment Flag}, \( F_{\text{obj}} \), verifies whether the solution is focused on solving the correct objective of the given question. \cite{anand2023gec} (ii) \textbf{Variables Application Flag}, \( F_{\text{val}} \), verifies whether the solution uses the correct values for all variables and constants provided in the question, ensuring their correct values are applied in formulae and reasoning.

\begin{figure*}[ht]
    \centering
    \includegraphics[width=15 cm]{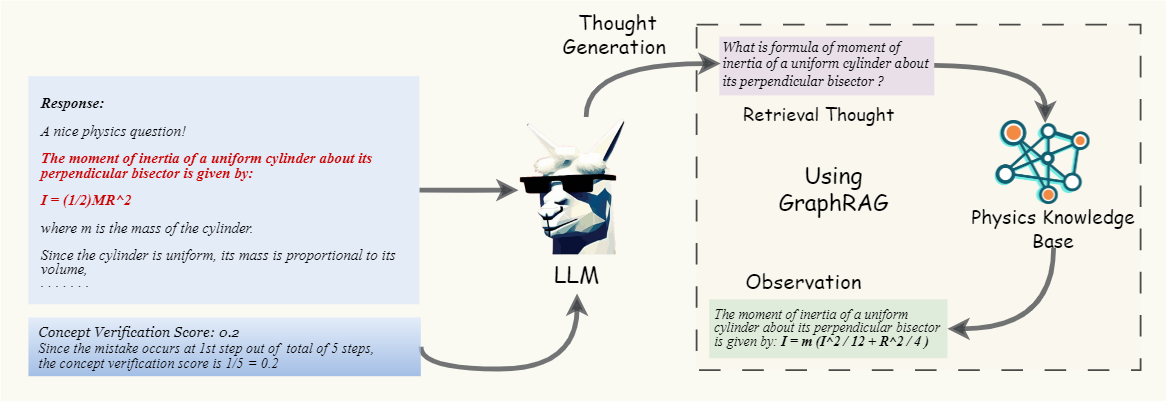} 
    \caption{The illustration of thought generation and concept retrieval for conceptual error refinement in LLM response. Given the response and concept verification score, LLM generates a retrieval thought, which acts as a query to retrieve the correct conceptual context from an physics knowledge base using GraphRAG.}
    \label{fig:pipeline_1}
\end{figure*}

\begin{figure*}[ht]
    \centering
    \includegraphics[width=15.5cm]{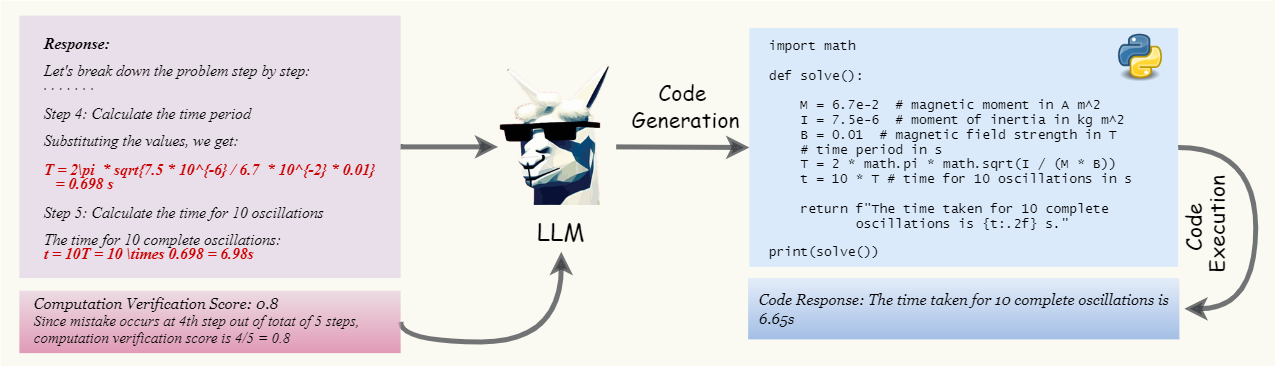} 
    \caption{The illustration of code generation and execution for computation error refinement in LLM response. Given the response and computation verification score, LLM generates a code to perform the correct required computation; the code is then executed to obtain the response. }
    \label{fig:pipeline_2}
\end{figure*}

\textbf{Concept Verification Score:} We instruct GPT-4o to check the given solution against the relevant concept and formulae required to solve the given problem, based on its own understanding of the question. A score (\( Score_{\text{concept}} \)) is assigned to each solution to quantify the correctness of the applied physics concepts and formulae. The score is designed to identify the stage at which any conceptual or formulae error first occurs, if at all. \( Score_{\text{concept}} \) ranges from 0 to 1, where a lower score indicates an earlier-stage error and a higher score indicates a later-stage error in the solution process. The score is calculated as follows:
\[
   Score_{\text{concept}} = 
   \begin{cases} 
   \frac{n}{N} & \text{if } 1 \leq n < N \text{ (error at step } n) \\
   \frac{n}{N+1} & \text{if } n = N \text{ (error at last step)} \\
   1 & \text{if no errors occur}
   \end{cases}
   \]
   where \( n \) is the step at which the first error occurs, and \( N \) is the total number of steps in the solution process.
   
\textbf{Computation Verification Score:} We employ GPT-4o with OpenAI Code Interpreter for generation and execution of python code to evaluate the correctness of all arithmetic and algebraic operations in the given solution. Similar to the \( Score_{\text{concept}} \), we assign \( Score_{\text{comp}} \) to each solution. This score quantifies the accuracy of the mathematical computation performed, ranges from 0 to 1, and is calculated similar to \( Score_{\text{concept}} \). All the computations are evaluated with an error tolerance of 0.1. Using Code Interpreter enables us to leverage the code generation capabilities of GPT-4o rather than solely relying on its mathematical reasoning, which sometimes can lead to erroneous scores. Recent works, such as \cite{zhou2023solving} and \cite{wang2023mathcoder}, highlights the remarkable capability of OpenAI Code Interpreter in solving challenging math problems and self-verification. \cite{anand2024mm}

We utilize GPT-4o solely for error identification, which guides the routing to the appropriate refinement agent. The scores are used as the feedback to help refinement agents understand the first stage of the mistake from where the refinement needs to be initiated. 

\subsection{Refinement Agents}
To address the three key errors in LLM generated solutions, we introduce a set of specialized refinement agents. Each agent is designed to rectify a specific type of error within the solution, ensuring targeted and effective corrections. The refinement agents use the same LLM with which the original solutions are generated.

\subsubsection{Miscomprehension Refinement}
Although there are very few cases of problem miscomprehension in LLM generated solutions, once identified these mistakes can easily be corrected with simple instruction prompting:

\textit{You are tasked with solving a physics problem. Here is the question: [question],
The following is your generated solution: [solution],
In the generated solution, the correct objective of the question is not being addressed. The solutions contains mistakes which leads to
misalignment with the objective of the question. Please carefully review the question \& understand the objective in detail and regenerate the solution accordingly.}

The above prompt assists in refining the solution to align with the correct objective of the given question. This may involve regenerating the entire solution or correcting an intermediate mistake to ensure the solution addressed the correct objective. Similarly, any incorrect variable values used within the solution is corrected using instruction prompting.

\subsubsection{Concept Refinement}
 To address the incorrect concepts and formulae applied in the LLM's solutions, we utilize an external physics knowledge base. This is necessary because LLMs may not always have access to or accurately retrieve the correct formulae, as this information may not be embedded in their internal knowledge. The conceptual refinement occurs in two steps:

1. \textbf{Error Identification \& Thought Generation:} Given a solution \( S_{\text{orig}} \) and a concept score \( Score_{\text{concept}} \), the LLM systematically reviews the solution to identify the earliest stage where an incorrect concept or formula has been applied. \( Score_{\text{concept}} \) pinpoints this stage of error within the solution. LLM then generates a retrieval thought \( T_{\text{R}} \) for the the concept or formula required at the failure stage. The thought is structured to be simple and sequential query.

2. \textbf{Concept Retrieval \& Solution Refinement:} Given the retrieval thought \( T_{\text{R}} \) and physics knowledge base \( K_{\text{P}} \), we use GraphRAG (Edge et al., 2024) to query the \( K_{\text{P}} \) to retrieve an observation \( O_{\text{T}} \), based on \( T_{\text{R}} \) as demonstrated in Figure \ref{fig:pipeline_1}. \( O_{\text{T}} \) contains the correct context for concept and formulae required at the stage of failure. The LLM then initiates the refinement from this stage using the information present in \( O_{\text{T}} \), resulting in the refined solution \( S_{\text{refined}} \) with corrected physics concepts and reasoning.

\subsubsection{Computational Refinement}
Inspired by recent works such as PAL \cite{gao2023pal}, PoT \cite{chen2022program}, CSV \cite{zhou2023solving}, MathCoder \cite{wang2023mathcoder}, we use code generation for the refinement of computational and mathematical errors within a solution. The computation score \( Score_{\text{comp}} \) allows the LLM to locate the first step of error and then initiate the refinement of the failure stage and subsequent computations. The process occurs in two steps:

1. \textbf{Code Generation \& Execution:} Given the original solution \( S_{\text{orig}} \) and computation score \( Score_{\text{comp}} \), the LLM first locates the error step and then generates a Python code \( C_{\text{p}} \) designed to accurately perform the necessary computation at the identified failure stage and produce the correct result. The generated code \( C_{\text{p}} \) is then executed to obtain the response \( R_{\text{c}} \) as shown in Figure \ref{fig:pipeline_2}.

2. \textbf{Solution Refinement:} The LLM is then instructed to refine \( S_{\text{orig}} \) using the correct response \( R_{\text{c}} \) generated by the code. This involves pinpointing the exact step where the error occurred, guided by the computation score \( Score_{\text{comp}} \), and integrating the correct computation from \( R_{\text{c}} \) into the solution. The refined solution \( S_{\text{refined}} \) is then presented with the corrected computations.

\begin{algorithm}
\small
\caption{Error Identification and Iterative Refinement}
\label{alg:algo1}
\begin{algorithmic}[1]
\REQUIRE Question $Q$, Initial Solution $S_0$,  GPT-4o $\mathcal{L}$, Refinement Agents $\mathcal{R}$, Maximum Iterations $N$, Threshold $\epsilon$
\ENSURE Final refined solution to $Q$
\STATE $i = 0$, $S_i = S_0$
\WHILE{$i < N$}
    \STATE $(F_{\text{obj}}^{i}, F_{\text{val}}^{i}, Score_{\text{concept}}^{i}, Score_{\text{comp}}^{i}) \leftarrow \mathcal{L}(Q, S_i)$
    
    \IF{$F_{\text{obj}}^i == -1$ \OR $F_{\text{val}}^i == -1$}
        \STATE $S_{i+1} \leftarrow \mathcal{R}_{\text{miscomprehension}}(Q, S_i)$
    \ELSIF{$Score_{\text{concept}}^i < 1 - \epsilon$}
        \STATE $S_{i+1} \leftarrow \mathcal{R}_{\text{concept}}(Q, S_i)$
    \ELSIF{$Score_{\text{comp}}^i < 1 - \epsilon$}
        \STATE $S_{i+1} \leftarrow \mathcal{R}_{\text{computation}}(Q, S_i)$
    \ELSE
        \RETURN $S_i$
    \ENDIF
    \STATE $i \leftarrow i + 1$
\ENDWHILE
\RETURN $S_N$

\end{algorithmic}
\end{algorithm}

\subsection{Agent Routing and Iterative Refinement}
After the error identification, the respective refinement agents are activated to mitigate these errors. The agent routing follows a prioritized sequence: 1.) miscomprehension refinement, 2.) concept refinement, 3.) computational refinement. This prioritization mirrors the human approach to solving physics problems: first, understanding the objectives and variables; next, identifying relevant concepts and formulae; and finally, applying them to perform the necessary computations.

The activated refinement agent then acts upon the solution to mitigate the error. The solution undergoes iterative cycles of error identification and refinement until all flags and scores are resolved or a maximum iteration limit is reached. This process ensures that all errors are corrected without introducing new ones in final refined solution. The complete process is illustrated in Algorithm~\ref{alg:algo1}.
\begin{table*}[ht]
\centering
\fontsize{9pt}{10pt}\selectfont
\setlength{\tabcolsep}{3pt} 
\begin{tabular}{l | c c c | c c c | c c c | c c c }
\toprule
\multirow{2}{*}{\textbf{Model}} & \multicolumn{3}{c}{\textbf{SciEval-Static}} & \multicolumn{3}{c}{\textbf{PhysicsQA}} & \multicolumn{3}{c}{\textbf{MMLU - High}} & \multicolumn{3}{c}{\textbf{MMLU - College }} \\ 
\cmidrule(lr){2-4} \cmidrule(lr){5-7} \cmidrule(lr){8-10} \cmidrule(lr){11-13} 
                                & \textbf{AO} & \textbf{CoT} & \textbf{3-Shot} & \textbf{AO} & \textbf{CoT} & \textbf{3-Shot} & \textbf{AO} & \textbf{CoT} & \textbf{3-Shot} & \textbf{AO} & \textbf{CoT} & \textbf{3-Shot} \\ \midrule
LLaMa-3-70B                     & 70.07\% & 82.23\% & 63.41\% & 38.37\% & 56.76\% & 59.29\% & 60.16\% & 72.88\% & 73.66\% & 59.41\% & 71.76\% & 71.76\% \\ 
LLaMa 3.1 405B                  & \textbf{79.87\%} & 89.63\% & \textbf{82.92\%} & \textbf{50.81\%} & 76.75\% & \textbf{78.37\%} & \textbf{72\%} & 91.52\% & \textbf{88.98\%} & \textbf{75.29\%} & \textbf{88.23\%} & \textbf{85.29\%} \\ 
Gemma-2-27B                     & 60.36\% & 79.26\% & 53.04\% & 39.18\% & 54.59\% & 59.45\% & 55.93\% & 77.11\% & 74.45\% & 51.11\% & 73.52\% & 67.64\% \\ 
Gemini 1.5 Flash                & 68.29\% & 85.97\% & 81.70\% & 44.86\% & 62.97\% & 69.72\% & 58.47\% & 79.66\% & 80.05\% & 60.58\% & 72.35\% & 72.94\% \\ 
GPT 3.5 Turbo                   & 41.46\% & 66.46\% & 48.78\% & 28.10\% & 42.70\% & 42.70\% & 47.45\% & 58.47\% & 33.89\% & 35.29\% & 50.58\% & 42.35\% \\ 
GPT4o                           & 64.02\% & \textbf{92.68}\% & 81.09\% & 49.45\% & \textbf{79.45\%} & \textbf{78.37\%} & 62.71\% & \textbf{94.06\%} & 87.28\% & 70\% & 84.70\% & 84.17\% \\ 
\bottomrule
\end{tabular}
\caption{Experimentation of Answer-Only (AO) , COT and Few-Shot (3-shot) on different Datasets}
\label{experimentalresults}
\end{table*}


\begin{table}[ht]
\centering
\fontsize{9pt}{10pt}\selectfont
\setlength{\tabcolsep}{2pt} 
\begin{tabular}{l|l|c|c|c|c}
\toprule
\textbf{Model} & \textbf{Dataset} & \textbf{AO} & \textbf{COT} & \textbf{3-Shot} & \textbf{MORA} \\
\midrule
\textbf{Gemma 2} & MMLU College &51.11\% & 73.52\% & 67.64\% & \textbf{82.20\%}  \\
 \textbf{27B} & MMLU High School & 55.93\% & 77.11\% & 74.45\% & \textbf{75.88\%} \\
 & PhysicsQA & 39.18\% & 54.59\% & 59.45\% & \textbf{70.62\%} \\
 & SciEval-Static & 60.36\% & 79.26\% & 53.04\% & \textbf{88.76\%} \\
\midrule
\textbf{LLaMa 3} & MMLU College & 59.41\% & 71.76\% & 71.76\% & \textbf{78.82\%} \\
\textbf{70B} & MMLU High School & 60.16\% & 72.88\% & 73.66\% & \textbf{78.81}\%  \\
 & PhysicsQA & 38.37\% & 56.76\% & 59.29\% & \textbf{70.14\%} \\
 & SciEval-Static & 70.07\% & 82.23\% & 63.41\% & \textbf{86.58\%} \\
\bottomrule
\end{tabular}
\caption{Comparison of baseline approaches with MoRA across four datasets: SciEval-Static, PhysicsQA, MMLU High School and College based on final answer accuracy.}
\label{experimentalresults2}
\end{table}

\section{Experiments}
\subsection{Setup}
\subsubsection{Datasets}
In our experiments, we use four datasets: SciEval-Static, PhysicsQA, MMLU High School and MMLU College. SciEval-Static is a subset of SciEVal \cite{sun2023scieval}, consisting 164 questions from physics divided into multiple sub-topics. MMLU \cite{hendrycks2021measuring}, consists of a 118 College level and 173 high school multiple-choice questions from various disciplines.

\subsubsection{LLMs}
We utilize the API of a range of models with varying parameters and capabilities including LLaMa-3-70B, LLaMa 3.1-405B, Gemma-2-27B, Gemini-1.5-Flash, GPT-3.5 Turbo and GPT-4 as our LLMs for the evaluation. We use same prompts for all the datasets and LLMs during our evaluation.

\subsubsection{Baselines}
We employ an Answer-only approach (AO), where the model is given a question with four options and asked to select the correct answer without any explanation  relying solely on its pre-existing knowledge . In contrast, few-shot prompting \cite{xu2023expertprompting, yasunaga2023large} uses a few examples to help the model learn and apply that knowledge to similar tasks. Chain-of-Thought (CoT) prompting \cite{wei2022chain} guides the model to generate intermediate reasoning steps, improving its performance on complex tasks by breaking them down into smaller, more manageable parts. These three approaches form our primary baselines.

\subsubsection{Evaluation}

Most of the existing works \cite{luo2023wizardmath} , \cite{chern2023generative} , \cite{yu2023metamath} measure the mathematical reasoning quality of LLMs by directly comparing the final answer and calculating the overall accuracy on a given dataset. \cite{anand2024mathify} We choose to follow the same evaluation for physics reasoning as well.

\subsection{Results}

In Table \ref{experimentalresults} we present results from our experiments reveal compelling insights into the strengths and challenges of various models across diverse benchmarks. In the SciEval-Static benchmark, LLaMa-3-70B and Gemini 1.5 Flash stand out, with LLaMa-3-70B achieving an accuracy of 82.23\% using (CoT), and Gemini 1.5 Flash not far behind at 85.97\%. In the PhysicsQA domain, which demands intricate reasoning skills, the models face more significant challenges. LLaMa-3-70B and Gemma 2-27B both show improved performance with CoT, reaching 56.76\% and 54.59\%, respectively. On the MMLU-High benchmark, LLaMa-3-70B continues to perform solidly, achieving 72.88\% with CoT, while Gemini 1.5 Flash pushes ahead to 79.66\%. Interestingly, in MMLU-College, a benchmark with a mix of academic and reasoning tasks, Gemma 2-27B shows a significant leap in performance with CoT, reaching 73.52\%, which surpasses its base score by over 22\%, indicating the effectiveness of CoT in enhancing reasoning in academic settings. 

As shown in Table \ref{experimentalresults2}, MoRA framework delivers marked improvements across all benchmarks for both LLaMA-3-70B and Gemma-2-27B models. In SciEval-Static, the introduction of MoRA enhances LLaMA-3-70B's accuracy from 82.23\% (CoT) to 86.58\%, and Gemma-2-27B sees a boost from 79.26\% (CoT) to 88.76\%. In the PhysicsQA dataset, MoRA significantly improves LLaMA-3-70B’s performance from 59.29\% (3-shot) to 70.14\%, and Gemma-2-27B's from 59.45\% (3-shot)to 70.62\%.  In MMLU High School, LLaMA-3-70B accuracy reaches from 73.66\% (3-Shot) to 78.81\% and in  Gemma-2-27B, accuracy reaches from 77.11\% (CoT) to 75.88\%. In MMLU College, LLaMA-3-70B accuracy reaches from 71.76\% (3-Shot) to 78.82\% and in  Gemma-2-27B, accuracy reaches from 73.52\% (CoT) to 82.20\%. The results show that even without extensive fine-tuning, these models can achieve competitive performance. These improvements demonstrate MoRA's ability to elevate smaller models to compete effectively with much larger ones across a range of complex tasks.

\section{Analysis}
In this section, we first conduct an in-depth error analysis of physics CoT solutions across various models and datasets, highlighting the error distribution that inspired the development of MoRA. We then present ablation studies, analyzing the effectiveness of each refinement agent in our framework.

\subsection{Error Analysis}

We perform manual analysis of the incorrect CoT solutions of GPT-4o, Llama-3-70B, and Gemma-2-27B on the following datasets: SciEval-Static, PhysicsQA, MMLU High School and College as shown in Table \ref{erroranalysis}. Based on this analysis here are our observations:

(i) \textbf{LLMs demonstrate good problem comprehension ability for physics question.} All models demonstrate strong problem comprehension across the datasets. GPT-4o excels, achieving near-perfect accuracy on SciEval-Static, MMLU College, and High School, with only minor errors in PhysicsQA. This suggests a deep understanding of physics problem structure. Llama-3-70B and Gemma-2-27B also perform well but show slightly higher error rates, particularly in PhysicsQA and SciEval-Static, indicating occasional missed details that need attention.

(ii) \textbf{Open source LLMs sometimes struggles to retrieve correct physics concept and formulae while reasoning.} On average, 18.11\% of questions in the PhysicsQA dataset are answered with conceptual errors by Gemma-2-27B and Llama-3-70B, highlighting the difficulty open-source LLMs face in applying correct concepts to physics problems. In contrast, GPT-4o excels with an average accuracy of 96.4\% across all four datasets. Notably, Gemma-2-27B outperforms Llama-3-70B on SciEval-Static, PhysicsQA, and MMLU High School. The high error rates of both Llama-3-70B and Gemma-2-27B on PhysicsQA suggest that medium-parameter open-source LLMs may still need external knowledge bases for complex physics problem-solving.

(iii) \textbf{Open-source LLMs struggles with algebraic and arithmetic computation required while solving physics questions.} On average, 21.62\% of questions in the PhysicsQA dataset are answered with computational mistakes by Gemma-2-27B and Llama-3-70B, highlighting challenges in executing correct calculations. GPT-4o excels with an average accuracy of 95.64\% across four datasets. Gemma-2-27B outperforms Llama-3-70B on SciEval Static and MMLU (High School and College), with both performing similarly on PhysicsQA. The accuracy gap between GPT-4o and Llama-3-70B (12.16\%) and GPT-4o and Gemma-2-27B (13.24\%) on PhysicsQA suggests that open-source LLMs could benefit from further refinement in handling complex calculations.


\begin{table}[ht]
\centering
\fontsize{9pt}{10pt}\selectfont
\setlength{\tabcolsep}{2pt} 
\begin{tabular}{l|l|c|c|c}
\toprule
\textbf{Error Type} & \textbf{Dataset} & \textbf{GPT-4o} & \textbf{Gemma} & \textbf{LLaMa} \\
& & & \textbf{2-27B}  & \textbf{3-70B}\\
\midrule
\textbf{Computational} & MMLU College & 2.54\% & 5.08\% & 9.32\% \\
 \textbf{Error} & MMLU High School & 2.35\% & 3.53\% & 6.47\% \\
 & PhysicsQA & 8.92\% & 22.16\% & 21.08\% \\
 & SciEval-Static & 3.06\% & 10.37\% & 10.98\% \\
\midrule
\textbf{Problem} & MMLU College & 0.00\% & 0.85\% & 1.69\% \\
\textbf{Miscomprehe-} & MMLU High School & 0.59\% & 0.59\% & 1.18\%  \\
 \textbf{nsion}& PhysicsQA & 0.54\% & 2.16\% & 1.62\% \\
 & SciEval-Static & 0.00\% & 1.22\% & 1.83\% \\
\midrule
\textbf{Wrong} & MMLU College & 0.85\% & 12.71\% & 10.17\% \\
\textbf{Concept} & MMLU High School & 3.53\% & 8.24\% & 11.76\% \\
 & PhysicsQA & 7.57\% & 17.11\% & 18.92\% \\
 & SciEval-Static & 1.22\% & 7.36\% & 9.15\% \\
\bottomrule
\end{tabular}
\caption{Error Analysis of incorrect physics CoT solutions of different models across four datasets.}
\label{erroranalysis}
\end{table}

\begin{table}[ht]
\centering
\fontsize{9pt}{10pt}\selectfont
\setlength{\tabcolsep}{3pt} 
\begin{tabular}{l|l|c|c}
\toprule
\textbf{Error Type} & \textbf{Dataset} & \textbf{Gemma}& \textbf{LLaMa}\\
& & \textbf{2-27B} & \textbf{3-70B} \\
\midrule
\textbf{Computational} & MMLU College & 100\% & 81.8\% \\
 \textbf{Refinement} & MMLU High School & 33.3\% & 75.0\%  \\
 & PhysicsQA & 73.3\% & 72.6\%   \\
 & SciEval-Static & 57.1\% & 60.0\%  \\
\midrule
\textbf{Miscomprehension} & MMLU College & 37.5\% & 33.3\%  \\
\textbf{Refinement} & MMLU High School & 16.7\% & 37.5\%   \\
 & PhysicsQA & 48.7\% & 46.9\%  \\
 & SciEval-Static & 62.5\% & 57.1\%  \\
\midrule
\textbf{Concept} & MMLU College & 100\% & 100\%  \\
 \textbf{Refinement} & MMLU High School & 100\% & 100\%  \\
 & PhysicsQA & 62.5\% & 66.7\%  \\
 & SciEval-Static & 100\% & 66.7\%  \\
\bottomrule
\end{tabular}
\caption{Ablation studies for different refinement agent in MoRA using Gemma-2-27B and Llama-3-70B across four datasets, evaluated by refinement rate.}
\label{ablation}
\end{table}
\subsection{Ablation}
To understand the effectiveness of each refinement agent, we conduct ablation of each of the refinement agents with Llama-3-70B and Gemma-2-70B in terms of their refinement rate across different datasets as shown in Table \ref{ablation}. Here are our observations:

(i) \textbf{Problem miscomprhension errors are mitigated with simple instruction prompting and error feedback.} Llama-3-70B and Gemma-2-27B demonstrate good miscomprehension error refinement with instruction prompting, particularly in the MMLU datasets (High School and College), where both models achieve a perfect 100\% refinement rate. Llama-3-70B outperforms Gemma-2-27B slightly on PhysicsQA, with a refinement rate of 66.7\% compared to Gemma-2-27B's 62.5\%. However, Gemma-2-27B excels in SciEval Static and MMLU (College and High School). The decent accuracy on PhysicsQA suggests that open-source LLMs sometimes fail to rectify their misinterpretations in complex physics problems.

(ii) \textbf{Open-source LLM performers moderately in identifying the conceptual mistake and retrieval thought generation.} Llama-3-70B shows a 46.9\% refinement rate in PhysicsQA and slightly improves in SciEval Static at 57.1\%, but struggles with MMLU datasets, achieving 37.5\% and 33.3\% refinement in High School and College, respectively. Gemma-2-27B has a similar 48.7\% refinement rate in PhysicsQA and performs better in SciEval Static at 62.5\%, but underperforms significantly on MMLU High School with a 16.7\% refinement rate, improving modestly to 37.5\% on MMLU College. These results suggest that open-source LLMs have difficulty generating relevant retrieval thoughts at the initial stage of failure.

(iii) \textbf{Using code-driven refinement significantly corrects the computational errors. }Llama-3-70B and Gemma-2-27B excel in refining computational errors, demonstrating the effectiveness of code generation and execution. Llama-3-70B shows consistent performance with a 72.6\% refinement rate on PhysicsQA and strong results across SciEval Static (60\%), MMLU High School (75\%), and MMLU College (81.8\%). Gemma-2-27B slightly outperforms in PhysicsQA at 73.3\% and achieves a 100\% refinement rate in MMLU College. However, Gemma-2-27B's performance is more variable, particularly in MMLU High School (33.3\%), indicating potential challenges in specific code generation scenarios.
\section{Conclusion}
In this work, we introduce MoRA, a novel agentic refinement framework designed to mitigate three critical errors commonly made by LLMs when solving complex physics problems. MoRA first leverages GPT-4o for error identification and score assignment, which are then subsequently used to guide the refinement agents. This process is done iteratively until all the errors in the solution are mitigated successfully. To ensure a comprehensive evaluation, we also curate our own dataset, PhysicsQA, which includes a diverse set of high school-level physics problems. Our experiments and in-depth analysis across multiple datasets demonstrate that MoRA significantly enhances the performance of Llama-3-70B and Gemma-2-27B across multiple datasets.

\bibliography{aaai25}

\begin{thebibliography}{42}
\providecommand{\natexlab}[1]{#1}

\bibitem[{Anand et~al.(2023{\natexlab{a}})Anand, Addala, Baghel, Goel, Hira, Gupta, and Shah}]{highdatasetavinash}
Anand, A.; Addala, K.; Baghel, K.; Goel, A.; Hira, M.; Gupta, R.; and Shah, R.~R. 2023{\natexlab{a}}.
\newblock Revolutionizing High School Physics Education: A Novel Dataset.
\newblock In Goyal, V.; Kumar, N.; Bhowmick, S.~S.; Goyal, P.; Goyal, N.; and Kumar, D., eds., \emph{Big Data and Artificial Intelligence}, 64--79. Cham: Springer Nature Switzerland.
\newblock ISBN 978-3-031-49601-1.

\bibitem[{Anand et~al.(2023{\natexlab{b}})Anand, Goel, Hira, Buldeo, Kumar, Verma, Gupta, and Shah}]{SciPhyRAG}
Anand, A.; Goel, A.; Hira, M.; Buldeo, S.; Kumar, J.; Verma, A.; Gupta, R.; and Shah, R.~R. 2023{\natexlab{b}}.
\newblock Sciphyrag-retrieval augmentation to improve llms on physics q \&a.
\newblock In \emph{International Conference on Big Data Analytics}, 50--63. Springer.

\bibitem[{Anand et~al.(2023{\natexlab{c}})Anand, Gupta, Prasad, Goel, Lal, Verma, and Shah}]{anand2023kg}
Anand, A.; Gupta, M.; Prasad, K.; Goel, U.; Lal, N.; Verma, A.; and Shah, R.~R. 2023{\natexlab{c}}.
\newblock KG-CTG: citation generation through knowledge graph-guided large language models.
\newblock In \emph{International Conference on Big Data Analytics}, 37--49. Springer.

\bibitem[{Anand et~al.(2024{\natexlab{a}})Anand, Gupta, Prasad, Singla, Sanjeev, Kumar, Shivam, and Shah}]{anand2024mathify}
Anand, A.; Gupta, M.; Prasad, K.; Singla, N.; Sanjeev, S.; Kumar, J.; Shivam, A.~R.; and Shah, R.~R. 2024{\natexlab{a}}.
\newblock Mathify: Evaluating Large Language Models on Mathematical Problem Solving Tasks.
\newblock \emph{arXiv preprint arXiv:2404.13099}.

\bibitem[{Anand et~al.(2023{\natexlab{d}})Anand, Jairath, Lal, Bangar, Sikka, Verma, Shah, and Satoh}]{anand2023gec}
Anand, A.; Jairath, A.; Lal, N.; Bangar, S.; Sikka, J.; Verma, A.; Shah, R.~R.; and Satoh, S. 2023{\natexlab{d}}.
\newblock GEC-DCL: Grammatical Error Correction Model with Dynamic Context Learning for Paragraphs and Scholarly Papers.
\newblock In \emph{International Conference on Big Data Analytics}, 95--110. Springer.

\bibitem[{Anand et~al.(2024{\natexlab{b}})Anand, Jaiswal, Dharmadhikari, Marathe, Popat, Mital, Nair, Prasad, Kumar, Verma et~al.}]{anand2024geovqa}
Anand, A.; Jaiswal, R.; Dharmadhikari, A.; Marathe, A.; Popat, H.; Mital, H.; Nair, A.~R.; Prasad, K.; Kumar, S.; Verma, A.; et~al. 2024{\natexlab{b}}.
\newblock GeoVQA: A Comprehensive Multimodal Geometry Dataset for Secondary Education.
\newblock In \emph{2024 IEEE 7th International Conference on Multimedia Information Processing and Retrieval (MIPR)}, 102--108. IEEE.

\bibitem[{Anand et~al.(2024{\natexlab{c}})Anand, Kapuriya, Kirtani, Singh, Saraf, Lal, Kumar, Shivam, Verma, Shah et~al.}]{anand2024PhyRLHF}
Anand, A.; Kapuriya, J.; Kirtani, C.; Singh, A.; Saraf, J.; Lal, N.; Kumar, J.; Shivam, A.~R.; Verma, A.; Shah, R.~R.; et~al. 2024{\natexlab{c}}.
\newblock MM-PhyRLHF: Reinforcement Learning Framework for Multimodal Physics Question-Answering.
\newblock \emph{arXiv preprint arXiv:2404.12926}.

\bibitem[{Anand et~al.(2024{\natexlab{d}})Anand, Kapuriya, Singh, Saraf, Lal, Verma, Gupta, and Shah}]{anand2024mm}
Anand, A.; Kapuriya, J.; Singh, A.; Saraf, J.; Lal, N.; Verma, A.; Gupta, R.; and Shah, R. 2024{\natexlab{d}}.
\newblock MM-PhyQA: Multimodal Physics Question-Answering with Multi-image CoT Prompting.
\newblock In \emph{Pacific-Asia Conference on Knowledge Discovery and Data Mining}, 53--64. Springer.

\bibitem[{Anand et~al.(2023{\natexlab{e}})Anand, Prasad, Goel, Gupta, Lal, Verma, and Shah}]{anand2023context}
Anand, A.; Prasad, K.; Goel, U.; Gupta, M.; Lal, N.; Verma, A.; and Shah, R.~R. 2023{\natexlab{e}}.
\newblock Context-enhanced language models for generating multi-paper citations.
\newblock In \emph{International Conference on Big Data Analytics}, 80--94. Springer.

\bibitem[{Arora, Singh et~al.(2023)}]{arora2023have}
Arora, D.; Singh, H.~G.; et~al. 2023.
\newblock Have llms advanced enough? a challenging problem solving benchmark for large language models.
\newblock \emph{arXiv preprint arXiv:2305.15074}.

\bibitem[{Brown et~al.(2020)Brown, Mann, Ryder, Subbiah, Kaplan, Dhariwal, Neelakantan, Shyam, Sastry, Askell et~al.}]{brown2020language}
Brown, T.; Mann, B.; Ryder, N.; Subbiah, M.; Kaplan, J.~D.; Dhariwal, P.; Neelakantan, A.; Shyam, P.; Sastry, G.; Askell, A.; et~al. 2020.
\newblock Language models are few-shot learners.
\newblock \emph{Advances in neural information processing systems}, 33: 1877--1901.

\bibitem[{Chen et~al.(2022)Chen, Ma, Wang, and Cohen}]{chen2022program}
Chen, W.; Ma, X.; Wang, X.; and Cohen, W.~W. 2022.
\newblock Program of thoughts prompting: Disentangling computation from reasoning for numerical reasoning tasks.
\newblock \emph{arXiv preprint arXiv:2211.12588}.

\bibitem[{Chern et~al.(2023)Chern, Zou, Li, Hu, Feng, Li, and Liu}]{chern2023generative}
Chern, E.; Zou, H.; Li, X.; Hu, J.; Feng, K.; Li, J.; and Liu, P. 2023.
\newblock Generative ai for math: Abel.
\newblock \emph{URL https://github. com/GAIR-NLP/abel}.

\bibitem[{Cobbe et~al.(2021)Cobbe, Kosaraju, Bavarian, Chen, Jun, Kaiser, Plappert, Tworek, Hilton, Nakano et~al.}]{cobbe2021training}
Cobbe, K.; Kosaraju, V.; Bavarian, M.; Chen, M.; Jun, H.; Kaiser, L.; Plappert, M.; Tworek, J.; Hilton, J.; Nakano, R.; et~al. 2021.
\newblock Training verifiers to solve math word problems.
\newblock \emph{arXiv preprint arXiv:2110.14168}.

\bibitem[{Edge et~al.(2024)Edge, Trinh, Cheng, Bradley, Chao, Mody, Truitt, and Larson}]{edge2024local}
Edge, D.; Trinh, H.; Cheng, N.; Bradley, J.; Chao, A.; Mody, A.; Truitt, S.; and Larson, J. 2024.
\newblock From local to global: A graph rag approach to query-focused summarization.
\newblock \emph{arXiv preprint arXiv:2404.16130}.

\bibitem[{Fu et~al.(2022)Fu, Peng, Sabharwal, Clark, and Khot}]{fu2022complexity}
Fu, Y.; Peng, H.; Sabharwal, A.; Clark, P.; and Khot, T. 2022.
\newblock Complexity-based prompting for multi-step reasoning.
\newblock In \emph{The Eleventh International Conference on Learning Representations}.

\bibitem[{Gao et~al.(2023)Gao, Madaan, Zhou, Alon, Liu, Yang, Callan, and Neubig}]{gao2023pal}
Gao, L.; Madaan, A.; Zhou, S.; Alon, U.; Liu, P.; Yang, Y.; Callan, J.; and Neubig, G. 2023.
\newblock Pal: Program-aided language models.
\newblock In \emph{International Conference on Machine Learning}, 10764--10799. PMLR.

\bibitem[{He et~al.(2024)He, Luo, Bai, Hu, Thai, Shen, Hu, Han, Huang, Zhang et~al.}]{he2024olympiadbench}
He, C.; Luo, R.; Bai, Y.; Hu, S.; Thai, Z.~L.; Shen, J.; Hu, J.; Han, X.; Huang, Y.; Zhang, Y.; et~al. 2024.
\newblock Olympiadbench: A challenging benchmark for promoting agi with olympiad-level bilingual multimodal scientific problems.
\newblock \emph{arXiv preprint arXiv:2402.14008}.

\bibitem[{Hendrycks et~al.(2021)Hendrycks, Burns, Kadavath, Arora, Basart, Tang, Song, and Steinhardt}]{hendrycks2021measuring}
Hendrycks, D.; Burns, C.; Kadavath, S.; Arora, A.; Basart, S.; Tang, E.; Song, D.; and Steinhardt, J. 2021.
\newblock Measuring mathematical problem solving with the math dataset.
\newblock \emph{arXiv preprint arXiv:2103.03874}.

\bibitem[{Lewis et~al.(2020)Lewis, Perez, Piktus, Petroni, Karpukhin, Goyal, K{\"u}ttler, Lewis, Yih, Rockt{\"a}schel et~al.}]{lewis2020retrieval}
Lewis, P.; Perez, E.; Piktus, A.; Petroni, F.; Karpukhin, V.; Goyal, N.; K{\"u}ttler, H.; Lewis, M.; Yih, W.-t.; Rockt{\"a}schel, T.; et~al. 2020.
\newblock Retrieval-augmented generation for knowledge-intensive nlp tasks.
\newblock \emph{Advances in Neural Information Processing Systems}, 33: 9459--9474.

\bibitem[{Li et~al.(2024)Li, Wang, Li, Guo, Zhang, and Feng}]{li2024evaluating}
Li, X.; Wang, W.; Li, M.; Guo, J.; Zhang, Y.; and Feng, F. 2024.
\newblock Evaluating Mathematical Reasoning of Large Language Models: A Focus on Error Identification and Correction.
\newblock \emph{arXiv preprint arXiv:2406.00755}.

\bibitem[{Ling et~al.(2024)Ling, Fang, Li, Huang, Lee, Memisevic, and Su}]{ling2024deductive}
Ling, Z.; Fang, Y.; Li, X.; Huang, Z.; Lee, M.; Memisevic, R.; and Su, H. 2024.
\newblock Deductive verification of chain-of-thought reasoning.
\newblock \emph{Advances in Neural Information Processing Systems}, 36.

\bibitem[{Lu et~al.(2022)Lu, Mishra, Xia, Qiu, Chang, Zhu, Tafjord, Clark, and Kalyan}]{lu2022learn}
Lu, P.; Mishra, S.; Xia, T.; Qiu, L.; Chang, K.-W.; Zhu, S.-C.; Tafjord, O.; Clark, P.; and Kalyan, A. 2022.
\newblock Learn to explain: Multimodal reasoning via thought chains for science question answering.
\newblock \emph{Advances in Neural Information Processing Systems}, 35: 2507--2521.

\bibitem[{Luo et~al.(2023)Luo, Sun, Xu, Zhao, Lou, Tao, Geng, Lin, Chen, and Zhang}]{luo2023wizardmath}
Luo, H.; Sun, Q.; Xu, C.; Zhao, P.; Lou, J.; Tao, C.; Geng, X.; Lin, Q.; Chen, S.; and Zhang, D. 2023.
\newblock Wizardmath: Empowering mathematical reasoning for large language models via reinforced evol-instruct.
\newblock \emph{arXiv preprint arXiv:2308.09583}.

\bibitem[{Ma et~al.(2024)Ma, Gou, Hao, Xu, Wang, Pan, Yang, Cao, and Sun}]{ma2024sciagent}
Ma, Y.; Gou, Z.; Hao, J.; Xu, R.; Wang, S.; Pan, L.; Yang, Y.; Cao, Y.; and Sun, A. 2024.
\newblock SciAgent: Tool-augmented Language Models for Scientific Reasoning.
\newblock \emph{arXiv preprint arXiv:2402.11451}.

\bibitem[{Miao, Teh, and Rainforth(2023)}]{miao2023selfcheck}
Miao, N.; Teh, Y.~W.; and Rainforth, T. 2023.
\newblock Selfcheck: Using llms to zero-shot check their own step-by-step reasoning.
\newblock \emph{arXiv preprint arXiv:2308.00436}.

\bibitem[{Mruthyunjaya et~al.(2023)Mruthyunjaya, Pezeshkpour, Hruschka, and Bhutani}]{mruthyunjaya2023rethinking}
Mruthyunjaya, V.; Pezeshkpour, P.; Hruschka, E.; and Bhutani, N. 2023.
\newblock Rethinking language models as symbolic knowledge graphs.
\newblock \emph{arXiv preprint arXiv:2308.13676}.

\bibitem[{Ouyang et~al.(2023)Ouyang, Zhang, Yan, Liu, Han, and Qin}]{ouyang2023structured}
Ouyang, S.; Zhang, Z.; Yan, B.; Liu, X.; Han, J.; and Qin, L. 2023.
\newblock Structured chemistry reasoning with large language models.
\newblock \emph{arXiv preprint arXiv:2311.09656}.

\bibitem[{Petroni et~al.(2020)Petroni, Piktus, Fan, Lewis, Yazdani, De~Cao, Thorne, Jernite, Karpukhin, Maillard et~al.}]{petroni2020kilt}
Petroni, F.; Piktus, A.; Fan, A.; Lewis, P.; Yazdani, M.; De~Cao, N.; Thorne, J.; Jernite, Y.; Karpukhin, V.; Maillard, J.; et~al. 2020.
\newblock KILT: a benchmark for knowledge intensive language tasks.
\newblock \emph{arXiv preprint arXiv:2009.02252}.

\bibitem[{Sun et~al.(2023{\natexlab{a}})Sun, Luo, Gong, Lin, Shen, Guo, and Duan}]{sun2023enhancing}
Sun, J.; Luo, Y.; Gong, Y.; Lin, C.; Shen, Y.; Guo, J.; and Duan, N. 2023{\natexlab{a}}.
\newblock Enhancing chain-of-thoughts prompting with iterative bootstrapping in large language models.
\newblock \emph{arXiv preprint arXiv:2304.11657}.

\bibitem[{Sun et~al.(2023{\natexlab{b}})Sun, Han, Zhao, Ma, Shen, Chen, Chen, and Yu}]{sun2023scieval}
Sun, L.; Han, Y.; Zhao, Z.; Ma, D.; Shen, Z.; Chen, B.; Chen, L.; and Yu, K. 2023{\natexlab{b}}.
\newblock SciEval: A Multi-Level Large Language Model Evaluation Benchmark for Scientific Research.
\newblock \emph{arXiv preprint arXiv:2308.13149}.

\bibitem[{Sun et~al.(2024)Sun, Han, Zhao, Ma, Shen, Chen, Chen, and Yu}]{sun2024scieval}
Sun, L.; Han, Y.; Zhao, Z.; Ma, D.; Shen, Z.; Chen, B.; Chen, L.; and Yu, K. 2024.
\newblock Scieval: A multi-level large language model evaluation benchmark for scientific research.
\newblock In \emph{Proceedings of the AAAI Conference on Artificial Intelligence}, volume~38, 19053--19061.

\bibitem[{Tyen et~al.(2024)Tyen, Mansoor, Carbune, Chen, and Mak}]{tyen-etal-2024-llms}
Tyen, G.; Mansoor, H.; Carbune, V.; Chen, P.; and Mak, T. 2024.
\newblock {LLM}s cannot find reasoning errors, but can correct them given the error location.
\newblock In Ku, L.-W.; Martins, A.; and Srikumar, V., eds., \emph{Findings of the Association for Computational Linguistics ACL 2024}, 13894--13908. Bangkok, Thailand and virtual meeting: Association for Computational Linguistics.

\bibitem[{Wang et~al.(2023)Wang, Ren, Zhou, Lu, Luo, Shi, Zhang, Song, Zhan, and Li}]{wang2023mathcoder}
Wang, K.; Ren, H.; Zhou, A.; Lu, Z.; Luo, S.; Shi, W.; Zhang, R.; Song, L.; Zhan, M.; and Li, H. 2023.
\newblock Mathcoder: Seamless code integration in llms for enhanced mathematical reasoning.
\newblock \emph{arXiv preprint arXiv:2310.03731}.

\bibitem[{Wei et~al.(2022)Wei, Wang, Schuurmans, Bosma, Xia, Chi, Le, Zhou et~al.}]{wei2022chain}
Wei, J.; Wang, X.; Schuurmans, D.; Bosma, M.; Xia, F.; Chi, E.; Le, Q.~V.; Zhou, D.; et~al. 2022.
\newblock Chain-of-thought prompting elicits reasoning in large language models.
\newblock \emph{Advances in neural information processing systems}, 35: 24824--24837.

\bibitem[{Xu et~al.(2023)Xu, Yang, Lin, Wang, Zhou, Zhang, and Mao}]{xu2023expertprompting}
Xu, B.; Yang, A.; Lin, J.; Wang, Q.; Zhou, C.; Zhang, Y.; and Mao, Z. 2023.
\newblock Expertprompting: Instructing large language models to be distinguished experts.
\newblock \emph{arXiv preprint arXiv:2305.14688}.

\bibitem[{Yao et~al.(2024)Yao, Yu, Zhao, Shafran, Griffiths, Cao, and Narasimhan}]{yao2024tree}
Yao, S.; Yu, D.; Zhao, J.; Shafran, I.; Griffiths, T.; Cao, Y.; and Narasimhan, K. 2024.
\newblock Tree of thoughts: Deliberate problem solving with large language models.
\newblock \emph{Advances in Neural Information Processing Systems}, 36.

\bibitem[{Yasunaga et~al.(2023)Yasunaga, Chen, Li, Pasupat, Leskovec, Liang, Chi, and Zhou}]{yasunaga2023large}
Yasunaga, M.; Chen, X.; Li, Y.; Pasupat, P.; Leskovec, J.; Liang, P.; Chi, E.~H.; and Zhou, D. 2023.
\newblock Large language models as analogical reasoners.
\newblock \emph{arXiv preprint arXiv:2310.01714}.

\bibitem[{Yu et~al.(2023)Yu, Jiang, Shi, Yu, Liu, Zhang, Kwok, Li, Weller, and Liu}]{yu2023metamath}
Yu, L.; Jiang, W.; Shi, H.; Yu, J.; Liu, Z.; Zhang, Y.; Kwok, J.~T.; Li, Z.; Weller, A.; and Liu, W. 2023.
\newblock Metamath: Bootstrap your own mathematical questions for large language models.
\newblock \emph{arXiv preprint arXiv:2309.12284}.

\bibitem[{Yuan et~al.(2024)Yuan, Yuan, Li, Dong, Lu, Tan, Zhou, and Zhou}]{yuan2024scaling}
Yuan, Z.; Yuan, H.; Li, C.; Dong, G.; Lu, K.; Tan, C.; Zhou, C.; and Zhou, J. 2024.
\newblock Scaling Relationship on Learning Mathematical Reasoning with Large Language Models.

\bibitem[{Zhang et~al.(2022)Zhang, Zhang, Li, and Smola}]{zhang2022automatic}
Zhang, Z.; Zhang, A.; Li, M.; and Smola, A. 2022.
\newblock Automatic chain of thought prompting in large language models.
\newblock \emph{arXiv preprint arXiv:2210.03493}.

\bibitem[{Zhou et~al.(2023)Zhou, Wang, Lu, Shi, Luo, Qin, Lu, Jia, Song, Zhan et~al.}]{zhou2023solving}
Zhou, A.; Wang, K.; Lu, Z.; Shi, W.; Luo, S.; Qin, Z.; Lu, S.; Jia, A.; Song, L.; Zhan, M.; et~al. 2023.
\newblock Solving challenging math word problems using gpt-4 code interpreter with code-based self-verification.
\newblock \emph{arXiv preprint arXiv:2308.07921}.

\end{thebibliography}

\end{document}